\documentclass[conference]{IEEEtran}
\IEEEoverridecommandlockouts

\usepackage{graphicx}
\usepackage{amsmath}
\usepackage{amssymb}
\usepackage{multirow}
\usepackage{booktabs}
\usepackage{cite}
\usepackage{url}
\usepackage{hyperref} 

\graphicspath{{./}{./figs/}}

\begin{document}

\title{Reconstruction-Driven Multimodal Representation Learning for Automated Media Understanding}

\author{
\IEEEauthorblockN{Yassir Benhammou, Suman Kalyan, Sujay Kumar}
\IEEEauthorblockA{NStarX Inc.\\
Ballwin, MO, USA \\
Email: \{yassir.benhammou, suman.kalyan, sujay\}@nstarxinc.com}
}

\maketitle

\begin{abstract}
Broadcast and media organizations increasingly rely on artificial intelligence to automate the labor-intensive processes of content indexing, tagging, and metadata generation. However, existing AI systems typically operate on a single modality—such as video, audio, or text—limiting their understanding of complex, cross-modal relationships in broadcast material. In this work, we propose a \textbf{Multimodal Autoencoder (MMAE)} that learns unified representations across \textbf{text, audio, and visual} data, enabling end-to-end automation of metadata extraction and semantic clustering. The model is trained on the recently introduced \textbf{LUMA dataset}, a fully aligned benchmark of multimodal triplets representative of real-world media content. By minimizing joint reconstruction losses across modalities, the MMAE discovers modality-invariant semantic structures without relying on large paired or contrastive datasets. We demonstrate significant improvements in clustering and alignment metrics (Silhouette, ARI, NMI) compared to linear baselines, indicating that reconstruction-based multimodal embeddings can serve as a foundation for scalable metadata generation and cross-modal retrieval in broadcast archives. These results highlight the potential of reconstruction-driven multimodal learning to enhance automation, searchability, and content management efficiency in modern broadcast workflows.
\end{abstract}

\begin{IEEEkeywords}
Multimodal learning, autoencoders, AI automation, metadata generation, broadcast archives, cross-modal retrieval, LUMA dataset.
\end{IEEEkeywords}

\section{Introduction}

Modern broadcast and media production environments generate massive volumes of multimodal data spanning visual content, audio, text, and metadata that must be efficiently indexed, searched, and repurposed across platforms~\cite{tan2023multimodalreview,jiang2022survey}. As content archives expand and delivery channels diversify, broadcasters increasingly rely on automation and AI-driven analytics to manage the complexity of multimodal data streams~\cite{li2023mediaml,zhang2023crossmodal,chen2022visionlanguage}. However, conventional metadata extraction pipelines often operate on isolated modalities, limiting their ability to capture the semantic relationships that connect what is seen, heard, and described~\cite{gupta2023broadcastai,gao2022unified}.

To enable intelligent workflows such as automatic content tagging, cross-modal retrieval, and semantic similarity search, broadcast systems require unified representations that encode visual, auditory, and textual cues in a shared embedding space~\cite{xu2023survey,muller2023audiofusion}. This integration remains challenging due to differences in feature structure, data scale, and noise across modalities~\cite{zhao2022deepmultimodal,huang2023robustfusion}. In practice, developing such models demands methods that are both computationally efficient and data-efficient, capable of learning from moderately sized, domain-specific datasets rather than massive web-scale corpora~\cite{shen2023lightclip,zhang2023efficientmultimodal}.

Large-scale contrastive learning frameworks such as CLIP~\cite{radford2021clip} and ALIGN~\cite{jia2021align} have demonstrated impressive performance in visual–language understanding by training on hundreds of millions of image–text pairs. Yet, these models rely on extensive paired supervision and computational resources, making them difficult to deploy in broadcast environments where data are heterogeneous and often not labeled at scale~\cite{fang2023survey,li2023foundational}. Their architecture and training regimes are also primarily optimized for general-purpose web imagery, rather than the structured and multimodal nature of professional media workflows~\cite{han2023mediaclip,rao2024multimodalvision}.

This work investigates a complementary approach grounded in reconstruction-based multimodal representation learning~\cite{suzuki2022survey,guo2023multimodalvae,wu2022multimodalautoencoder}. Instead of contrasting matched and mismatched pairs, our framework learns a shared latent representation by jointly reconstructing each modality from a common embedding space. This method allows the model to capture intrinsic semantic relationships without relying on explicit negative sampling or large-scale supervision~\cite{he2022mae,shi2019mmvae}. We propose a \textbf{Multimodal Autoencoder (MMAE)} that integrates image, audio, and text modalities within a unified latent structure, trained to minimize cross-modal reconstruction losses~\cite{ngiam2011multimodal,suzuki2016jmvae}.

The MMAE is evaluated on the recently introduced \textbf{LUMA dataset}~\cite{luma2024}, which provides fully aligned triplets of images, audio clips, and captions designed for research on multimodal alignment and fusion. Using both quantitative (Silhouette, ARI, NMI) and qualitative (t-SNE, UMAP) analyses, we show that the MMAE achieves superior multimodal coherence and semantic clustering compared to linear baselines. These findings highlight the potential of reconstruction-based models to deliver interpretable, data-efficient multimodal representations suitable for next-generation broadcast automation systems, where interpretability, reproducibility, and integration efficiency are paramount~\cite{zhou2024broadcastai,wang2023multimodalframework}.

The remainder of this paper is organized as follows. Section~\ref{sec:relatedwork} reviews related research in multimodal representation learning and outlines the evolution of contrastive and generative approaches. Section~\ref{sec:dataset} introduces the LUMA dataset and describes the preprocessing pipeline used to ensure reproducibility. Section~\ref{sec:methodology} details the proposed Multimodal Autoencoder (MMAE) architecture and experimental setup. Section~\ref{sec:results} presents and discusses the quantitative and qualitative evaluation results. Finally, Section~\ref{sec:conclusion} concludes the paper and outlines directions for future work.

\section{Related Work}
\label{sec:relatedwork}
Multimodal representation learning has become a cornerstone of modern artificial intelligence, with applications ranging from cross-modal retrieval to content recommendation and broadcast automation~\cite{xu2023survey,jiang2022survey}. The objective is to learn embeddings that capture shared semantic information across diverse modalities—such as video, audio, and text—while preserving modality-specific details crucial for downstream understanding~\cite{zhao2022deepmultimodal,li2023foundational}. In broadcast and media contexts, these representations can automate content indexing, generate descriptive metadata, and align heterogeneous data sources for advanced search and recommendation systems~\cite{gupta2023broadcastai,li2023mediaml}.

Early multimodal fusion methods relied on simple concatenation or canonical correlation analysis (CCA)~\cite{hotelling1936cca}, which were unable to model nonlinear dependencies between modalities. Classical statistical techniques such as PCA~\cite{jolliffe2002pca} and K-Means~\cite{lloyd1982kmeans} provided interpretable baselines but lacked the representational depth required to encode high-level cross-modal semantics. With the rise of deep learning, models began to learn shared latent spaces capable of aligning disparate modalities in a data-driven manner.

A major breakthrough emerged with large-scale contrastive learning frameworks such as CLIP~\cite{radford2021clip} and ALIGN~\cite{jia2021align}, which optimize similarity between matched image–text pairs and dissimilarity for unmatched pairs. These architectures demonstrated remarkable zero-shot generalization and laid the groundwork for vision–language systems used in media tagging and retrieval~\cite{fang2023survey,han2023mediaclip}. Successors such as ALBEF~\cite{li2021albef}, LXMERT~\cite{tan2019lxmert}, and BLIP~\cite{li2022blip,li2023blip2} enhanced performance through attention mechanisms and pretraining strategies that combine caption generation with contrastive objectives. Nevertheless, their heavy reliance on web-scale paired data and supervised objectives makes them challenging to adapt to domain-specific environments like broadcasting, where data availability and compute capacity are limited~\cite{shen2023lightclip,zhang2023efficientmultimodal}.

In contrast, \textit{generative and reconstruction-based approaches} align modalities by learning to jointly reconstruct them from a shared latent representation. The early multimodal autoencoder proposed by Ngiam et al.~\cite{ngiam2011multimodal} demonstrated that modality-specific encoders and decoders with a common bottleneck can capture meaningful cross-modal relationships. This idea evolved into probabilistic frameworks such as JMVAE~\cite{suzuki2016jmvae}, MVAE~\cite{wu2018mvae}, and MMVAE~\cite{shi2019mmvae}, which enable flexible inference even when certain modalities are missing. More recent developments include masked autoencoders (MAE)~\cite{he2022mae}, which scale self-supervised reconstruction to large visual corpora, and self-supervised encoders for audio such as wav2vec 2.0~\cite{baevski2020wav2vec2}, HuBERT~\cite{hsu2021hubert}, and PANNs~\cite{kong2020panns}, as well as text encoders like BERT~\cite{devlin2019bert} that provide strong semantic grounding.

Recent works such as UniModal~\cite{wang2023unimodal}, BridgeTower~\cite{xu2022bridgetower}, and VATT~\cite{akbari2021vatt} extend multimodal learning to video–audio–text alignment, while other studies explore unified latent diffusion~\cite{rombach2022latentdiffusion} and multimodal generative pretraining~\cite{bao2022beit3}. Comprehensive surveys such as Suzuki’s review of deep multimodal generative models~\cite{suzuki2022survey} and Tan’s overview of multimodal fusion~\cite{tan2023multimodalreview} emphasize the advantages of reconstruction-driven approaches in discovering shared semantic manifolds, even when training data are limited or imperfectly aligned. In this context, the LUMA dataset~\cite{luma2024} serves as an ideal benchmark: its balanced, fully aligned triplets of images, audio, and text enable controlled studies of joint reconstruction and multimodal fusion.

Building upon these foundations, this work introduces a Multimodal Autoencoder (MMAE) tailored for discovering modality-invariant embeddings through deterministic joint reconstruction. Each modality is processed by a dedicated encoder–decoder pair that converges into a shared latent bottleneck~\cite{wu2022multimodalautoencoder,guo2023multimodalvae}. The resulting representation captures cross-modal semantics without contrastive supervision, offering a lightweight and interpretable alternative to data-intensive models. For broadcast and media AI systems, such an approach supports automated, scalable content understanding—enabling accurate cross-modal alignment while maintaining the transparency and reproducibility critical in professional media applications~\cite{zhou2024broadcastai,rao2024multimodalvision}.

\section{Dataset: The LUMA Benchmark}
\label{sec:dataset}

The experiments in this study are conducted using the \textbf{LUMA dataset}~\cite{luma2024}, a recently introduced benchmark designed for research on \textit{multimodal alignment, fusion, and representation learning}. Although originally developed for general AI research, LUMA reflects many characteristics of broadcast and media data—where visual, auditory, and textual cues co-occur naturally, as in video segments with corresponding audio tracks and subtitles. This makes it an appropriate testbed for evaluating models aimed at automation tasks such as cross-modal tagging, retrieval, and metadata enrichment.

\subsection{Dataset Composition}
Each sample in LUMA corresponds to a triplet $(x_I, x_A, x_T)$ comprising:
\begin{itemize}
    \item \textbf{Image ($x_I$):} a visual depiction of a scene or concept, analogous to a keyframe extracted from a video segment.
    \item \textbf{Audio ($x_A$):} a short human-recorded sound clip or spoken caption describing the same concept.
    \item \textbf{Text ($x_T$):} a concise natural-language caption semantically aligned with both the image and the audio.
\end{itemize}
All triplets are grouped into 50 semantic classes (e.g., ``airplane,'' ``dog,'' ``violin''), each representing a distinct concept shared across modalities—paralleling thematic categories common in media archives.

The dataset is divided into:
\begin{itemize}
    \item Training set: 21,000 aligned triplets (420 per class);
    \item Test set: 4,200 aligned triplets (84 per class);
    \item Out-of-distribution (OOD) set: 3,859 triplets from unseen domains and recording conditions.
\end{itemize}
This balanced composition ensures consistent per-class representation and standardized evaluation of multimodal generalization.

\begin{table}[htbp]
\centering
\caption{LUMA Dataset Statistics and Feature Dimensions}
\begin{tabular}{lccc}
\toprule
\textbf{Modality} & \textbf{Feature Dim.} & \textbf{Train} & \textbf{Test / OOD} \\
\midrule
Image & 50 & 21,000 & 4,200 / 3,859 \\
Audio & 1,024 & 21,000 & 4,200 / 3,859 \\
Text  & 768 & 21,000 & 4,200 / 3,859 \\
\bottomrule
\end{tabular}
\label{tab:luma_stats}
\end{table}

\subsection{Preprocessing and Feature Extraction}
To ensure reproducibility and computational efficiency, all raw data were transformed into compact feature embeddings using pretrained models widely adopted in AI and media analysis pipelines:
\begin{itemize}
    \item \textbf{Images:} extracted using a VGG11-BN network pretrained on ImageNet. The activations from the penultimate layer were projected to 50 dimensions via PCA to capture dominant semantic features.
    \item \textbf{Audio:} processed with a pretrained PANNs encoder producing 1,024-dimensional embeddings. Each clip was resampled to 16\,kHz, normalized, and converted to a mel-spectrogram before feature extraction.
    \item \textbf{Text:} encoded using the [CLS] embedding from a BERT-base model (12 layers, 768 hidden units). Captions were tokenized with WordPiece and truncated to 32 tokens for uniformity.
\end{itemize}

All feature vectors were standardized (zero mean, unit variance) and stored as NumPy arrays. Each file contains one aligned feature vector per sample, corresponding to the same semantic entity across modalities.

\subsection{Alignment and Reproducibility}
A defining characteristic of LUMA is its row-level alignment: each sample index $i$ refers to the same concept in all three modalities. This property closely mirrors real-world media assets, such as synchronized audio–subtitle–frame triplets in broadcast archives, and allows fully reproducible multimodal training without additional matching or synchronization steps.

The features are built as follows:
\begin{itemize}
    \item Images (50D). We use a VGG11-BN backbone (ImageNet pretrained) with a 50-way classification head. We take the \emph{50D logits} as features (no PCA).%
    \item Audio (1{,}024D). We extract embeddings with \emph{Wav2Vec2-Large} (torchaudio pipeline), mean-pooled over time to 1{,}024D. All clips are resampled to 16\,kHz and loudness-normalized before inference.
    \item Text (768D). We encode captions with \emph{BERT-base}, using the [CLS] token embedding (768D). Captions are tokenized with WordPiece and truncated to 32 tokens.
\end{itemize}

To promote transparent and reproducible experimentation while reducing computational requirements, we provide the complete set of \emph{pre-extracted, aligned features} for all data splits (Train/Test/OOD) in a single Google Drive bundle available \href{https://drive.google.com/file/d/1e27hZOsUlCFHTqUUmfDjp-e_EqkSUwD6/view?usp=sharing}{here}. 
A detailed \texttt{README} file included in the bundle describes the dataset structure, file contents, and step-by-step instructions for verifying data integrity and preparing the inputs used in all experiments.

\subsection{Qualitative Overview}
\begin{figure*}[htbp]
  \centering
  \includegraphics[width=\textwidth]{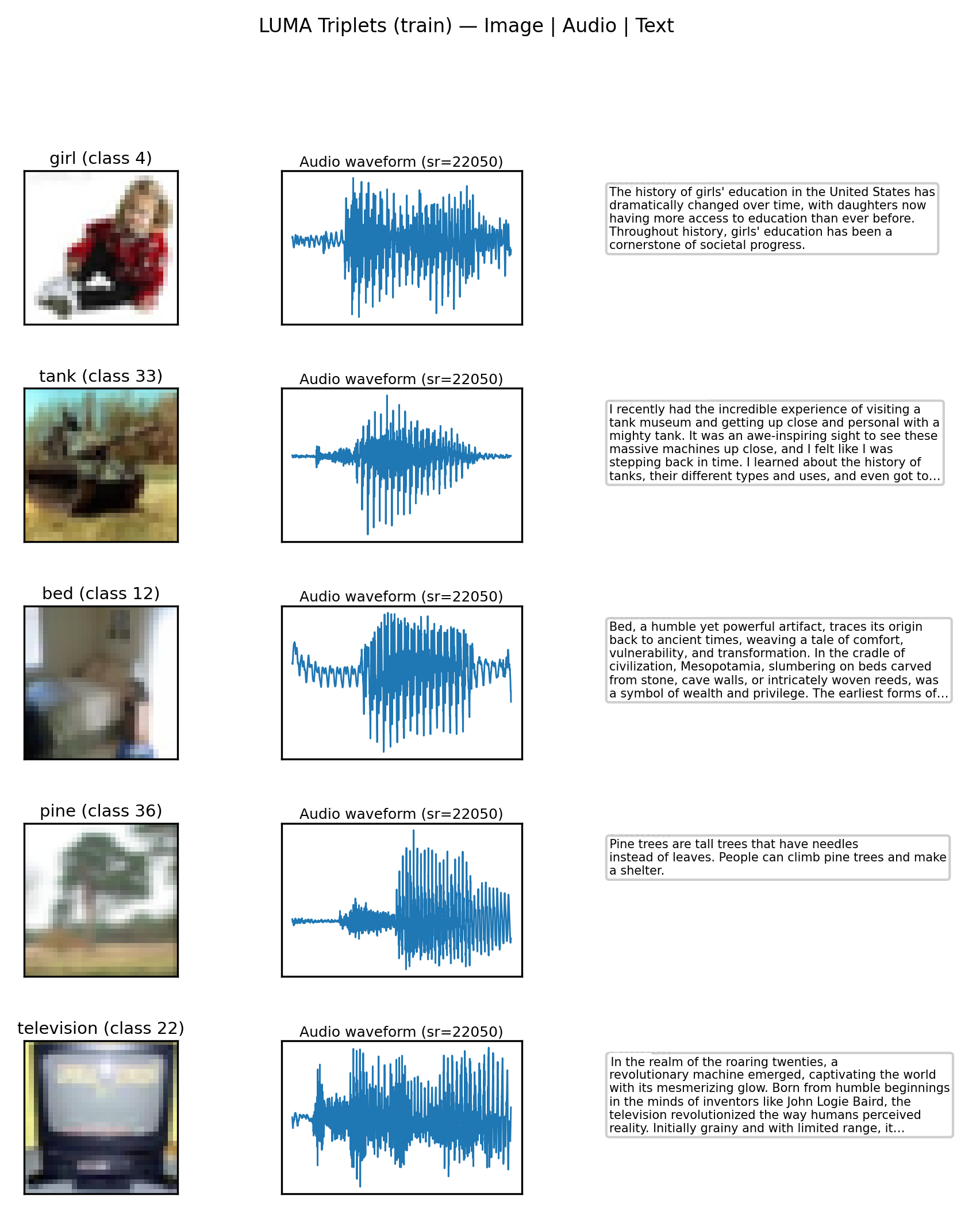}
  \caption{Aligned LUMA triplets (image, audio waveform, caption).
  Each row shows one example, with strict alignment between the three modalities.
  Left: visual depiction and class label; Middle: corresponding audio waveform; Right: natural-language caption. 
  The dataset’s structure parallels real-world audiovisual metadata—providing a realistic foundation for training reconstruction-based multimodal AI models for content understanding and automation.}
  \label{fig:luma_triplets}
\end{figure*}

Fig.~\ref{fig:luma_triplets} illustrates representative triplets from LUMA (image, audio waveform, caption). The figure highlights (i) precise triplet alignment, (ii) intra-class diversity across modalities, and (iii) the complementary nature of textual grounding. These properties make LUMA an ideal benchmark for evaluating multimodal AI systems designed for media automation, where semantically consistent alignment across modalities is essential.

\subsection{Why LUMA?}
Unlike large-scale contrastive datasets such as LAION or AudioCaps, LUMA offers a balanced, interpretable, and fully aligned multimodal structure at a moderate scale. This makes it especially valuable for research on autoencoder-based architectures, which depend on exact cross-modal correspondence for joint reconstruction learning. Its inclusion of an out-of-distribution (OOD) split further enables robust testing of generalization, a key requirement in operational broadcast systems.

In summary, LUMA provides a semantically rich and carefully aligned benchmark bridging the gap between synthetic alignment corpora and large-scale web datasets. While the dataset itself is not publicly distributed, its structure, feature specifications, and open-source experimental pipeline ensure complete methodological transparency and reproducibility.

\section{Methodology}
\label{sec:methodology} 
\begin{figure}[htbp]
    \centering
    \includegraphics[width=0.48\textwidth]{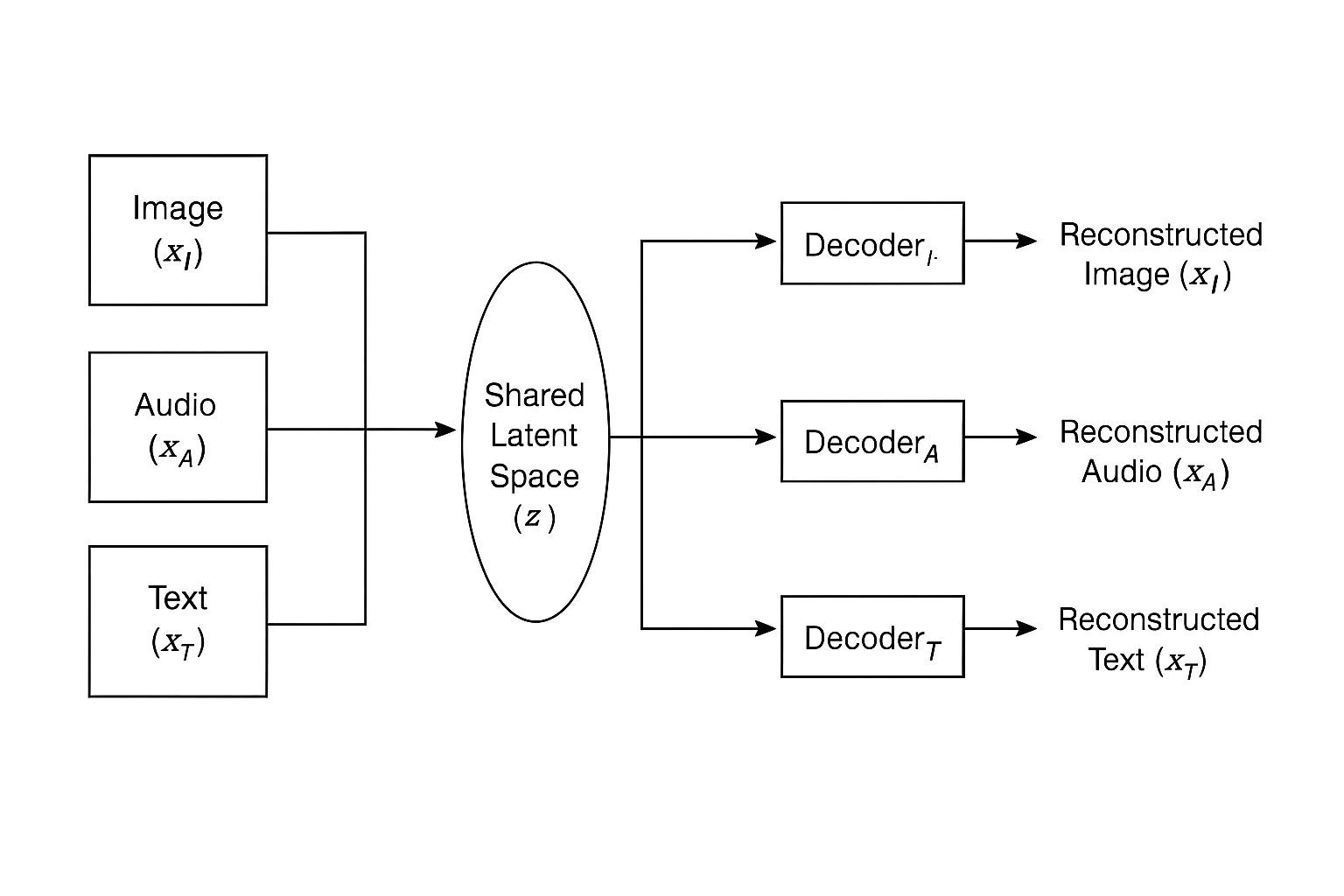}
    \caption{Architecture of the proposed Multimodal Autoencoder (MMAE). 
The model consists of three modality-specific encoders for image, audio, and text inputs, respectively. Each encoder maps its modality 
to a shared latent representation $z$, which captures modality-invariant semantic 
features. From this latent vector, three corresponding decoders $Decoder_I$, $Decoder_A$, and 
$Decoder_T$ reconstruct each modality, enforcing cross-modal consistency through 
joint reconstruction losses. This design encourages the shared latent space 
to align semantically equivalent content across modalities while preserving 
their unique characteristics.}
    \label{fig:mmae_architecture}
\end{figure}

This section details the different methodological stages of our study, from the design of baseline models to the implementation of the proposed Multimodal Autoencoder (MMAE). The overall pipeline is illustrated in Fig.~\ref{fig:mmae_architecture}.

\subsection{Baseline Methods}

\textbf{Single-modal clustering:}  
To establish baseline performance, each modality (image, audio, text) was independently projected using Principal Component Analysis (PCA) to its optimal reduced dimension—50 for images, 256 for audio, and 256 for text. The projected embeddings were then clustered using K-Means with $k \in \{30, 40, 42, 50, 60\}$. This allowed us to assess how well each modality captures semantic class structure in isolation.

\textbf{Fusion-based PCA:}  
As a linear multimodal baseline, feature matrices were concatenated and reweighted by coefficients $(\alpha, \beta, \gamma)$ corresponding to image, audio, and text modalities, respectively. The fused representation was projected via PCA to 50 dimensions and clustered using K-Means. This setup allows quantitative comparison between simple linear fusion and our proposed nonlinear autoencoder.

\subsection{Multimodal Autoencoder (MMAE)}

The proposed \textbf{MMAE} learns to encode each modality into a shared latent representation through reconstruction. Each modality $m \in \{I, A, T\}$ has an encoder $E_m(\cdot)$ and decoder $D_m(\cdot)$. The encoders map modality-specific inputs into a shared latent vector $z \in \mathbb{R}^{128}$:

\begin{equation}
z = E_I(x_I) = E_A(x_A) = E_T(x_T)
\end{equation}

Each decoder attempts to reconstruct its corresponding modality from $z$:
\begin{equation}
\hat{x}_m = D_m(z)
\end{equation}

The training objective minimizes the total reconstruction loss:
\begin{equation}
\mathcal{L}_{\text{rec}} = 
\|x_I - \hat{x}_I\|^2 + 
\|x_A - \hat{x}_A\|^2 + 
\|x_T - \hat{x}_T\|^2
\end{equation}

This joint objective forces the latent representation to retain shared semantic information across all modalities. During training, gradients are propagated jointly across encoders and decoders, enforcing a single, modality-invariant latent structure.

\subsection{Architecture Details}

Each encoder and decoder is implemented as a three-layer fully connected network with ReLU activations and batch normalization. For instance, the image encoder takes a 50-dimensional vector as input and outputs a 128-dimensional latent embedding via layers of sizes [128, 128, 128]. Audio and text encoders follow the same structure, starting from 1024 and 768 input dimensions respectively. Decoders mirror the encoder architecture symmetrically.

The shared latent bottleneck has a dimensionality of 128, chosen empirically to balance expressiveness and regularization. This bottleneck serves as the cross-modal embedding space for all experiments, including clustering and visualization.

\subsection{Training Setup}

All models were implemented in \texttt{PyTorch 2.3} using the \texttt{MPS} backend on macOS and verified on GPU-based systems. Training used the following configuration:

\begin{itemize}
    \item Optimizer: Adam
    \item Learning rate: $1 \times 10^{-3}$
    \item Batch size: 128
    \item Epochs: 100
    \item Loss: Mean Squared Error (MSE)
    \item Random seed: 42
\end{itemize}

The model converged stably within 50 epochs, with diminishing reconstruction loss observed thereafter. The final trained model weights were saved as \texttt{mmae\_z128.pt}, and latent embeddings were extracted for the train, test, and OOD splits as NumPy files for subsequent evaluation.

\subsection{Evaluation Protocol}

To evaluate the learned latent representations, we used three clustering metrics that quantify intra-cluster compactness and inter-cluster separability:
\begin{itemize}
    \item Silhouette Coefficient (Sil) – measures average similarity within clusters.
    \item Adjusted Rand Index (ARI) – compares cluster assignments to ground-truth labels.
    \item Normalized Mutual Information (NMI) – quantifies agreement between predicted and true partitions.
\end{itemize}

We additionally visualized the 2D projections of latent spaces using t-SNE and UMAP to assess the geometric coherence of multimodal embeddings qualitatively.

All experiment scripts used in this study are released together with the data in the Google Drive bundle available \href{https://drive.google.com/file/d/1e27hZOsUlCFHTqUUmfDjp-e_EqkSUwD6/view?usp=sharing}{here}. 
The accompanying \texttt{README} file provides detailed instructions on how to set up the environment, load the data, and reproduce every experiment reported in this paper.

\section{Results and Discussion}
\label{sec:results}
We evaluated the proposed Multimodal Autoencoder (MMAE) against PCA-based and single-modality baselines using clustering and visualization analyses on the LUMA test split. Performance was measured using the \textit{Silhouette Coefficient (Sil)}, \textit{Adjusted Rand Index (ARI)}, and \textit{Normalized Mutual Information (NMI)}, quantifying intra-class compactness, inter-class separability, and alignment with semantic labels, respectively. These metrics provide complementary insight into how well the learned embeddings capture shared structure across modalities—a key requirement for automated metadata generation and cross-modal retrieval systems in broadcast applications.

\subsection{Quantitative Evaluation}
Table~\ref{tab:singlemod} summarizes the results for single-modality clustering. As expected, the image modality yields the strongest discriminative structure, while text and audio show weaker class separability due to their higher abstraction and noise variability. This confirms that visual features tend to encode more consistent semantics than unimodal textual or acoustic embeddings.

Table~\ref{tab:fusion} reports the results for linear fusion baselines, in which weighted combinations of image, audio, and text features were projected using PCA before clustering. Although moderate improvements are observed relative to unimodal baselines, linear fusion remains limited in modeling complex intermodal dependencies, underscoring the need for nonlinear architectures that can capture semantic correspondences beyond simple feature concatenation.

The proposed MMAE model outperforms all baselines across all evaluation metrics, as shown in Table~\ref{tab:mmae}. At $k=42$, the model achieves its highest scores (Silhouette = 0.63, ARI = 0.91, NMI = 0.96), indicating strong agreement between learned clusters and ground-truth semantic categories. These results demonstrate that joint reconstruction learning successfully captures shared, modality-invariant representations, yielding embeddings that align closely with human-level semantic groupings.

\begin{table}[htbp]
\centering
\caption{Single-Modality Clustering Results (Test Set)}
\begin{tabular}{lccc}
\toprule
Modality & Sil & ARI & NMI \\
\midrule
Image & 0.418 & 0.545 & 0.724 \\
Text  & 0.147 & 0.091 & 0.380 \\
Audio & 0.087 & 0.053 & 0.227 \\
\bottomrule
\end{tabular}
\label{tab:singlemod}
\end{table}

\begin{table}[htbp]
\centering
\caption{Fusion K-Means Grid Search (Test Set)}
\begin{tabular}{cccc}
\toprule
k & Sil & ARI & NMI \\
\midrule
30 & 0.392 & 0.456 & 0.695 \\
42 & 0.418 & 0.545 & 0.724 \\
50 & 0.416 & 0.549 & 0.724 \\
\bottomrule
\end{tabular}
\label{tab:fusion}
\end{table}

\begin{table}[htbp]
\centering
\caption{MMAE Clustering Results (Test Latents, $z=128$)}
\begin{tabular}{cccc}
\toprule
k & Sil & ARI & NMI \\
\midrule
30 & 0.445 & 0.547 & 0.888 \\
40 & 0.620 & 0.826 & 0.949 \\
42 & \textbf{0.630} & \textbf{0.914} & \textbf{0.962} \\
50 & 0.586 & 0.896 & 0.948 \\
\bottomrule
\end{tabular}
\label{tab:mmae}
\end{table}

\subsection{Qualitative Analysis}

\begin{figure}[htbp]
    \centering
    \includegraphics[width=0.48\textwidth]{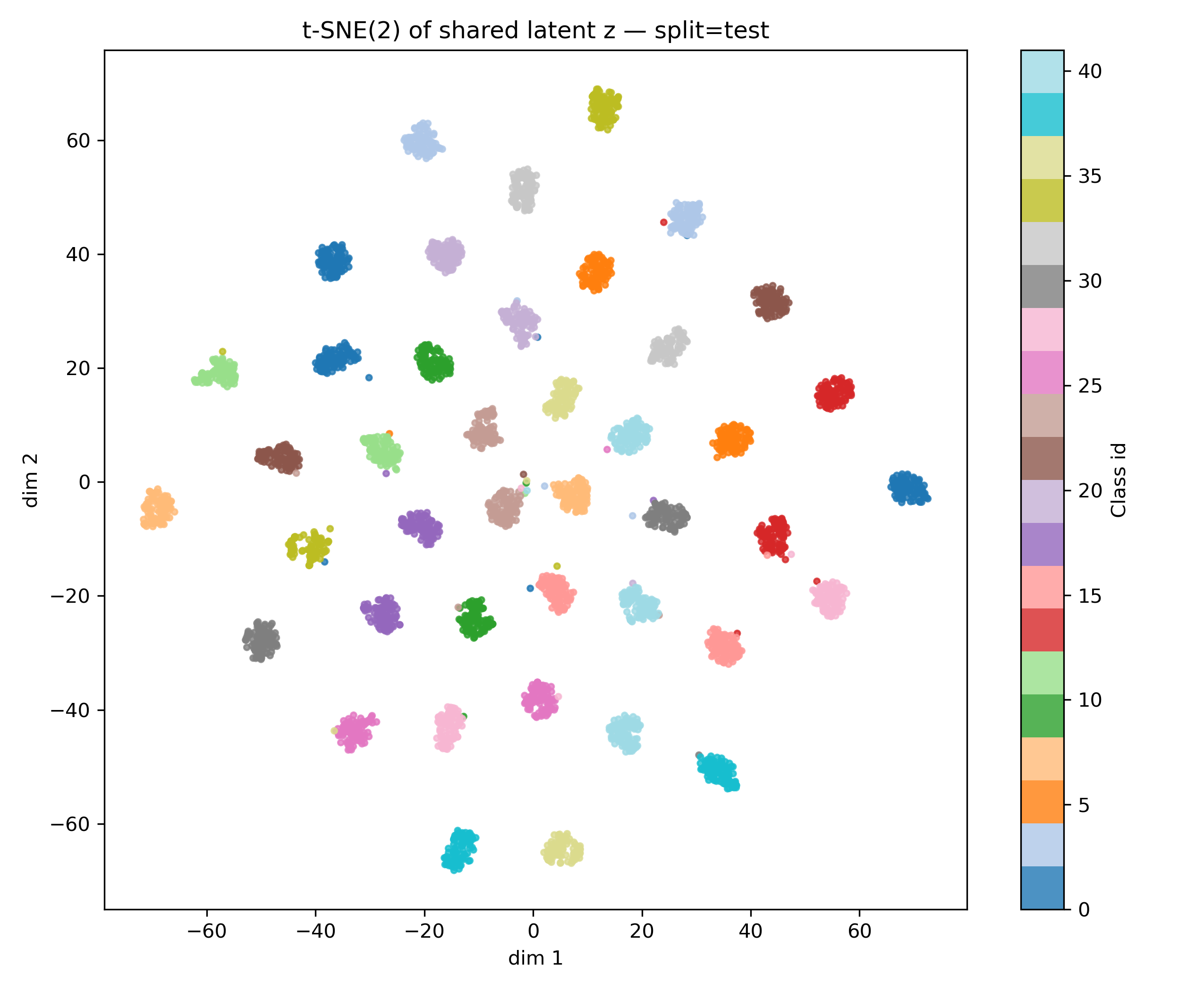}
    \caption{t-SNE projection of the MMAE latent space ($z=128$). Distinct, compact clusters reflect strong semantic alignment across modalities.}
    \label{fig:tsne_latent}
\end{figure}

\begin{figure}[htbp]
    \centering
    \includegraphics[width=0.48\textwidth]{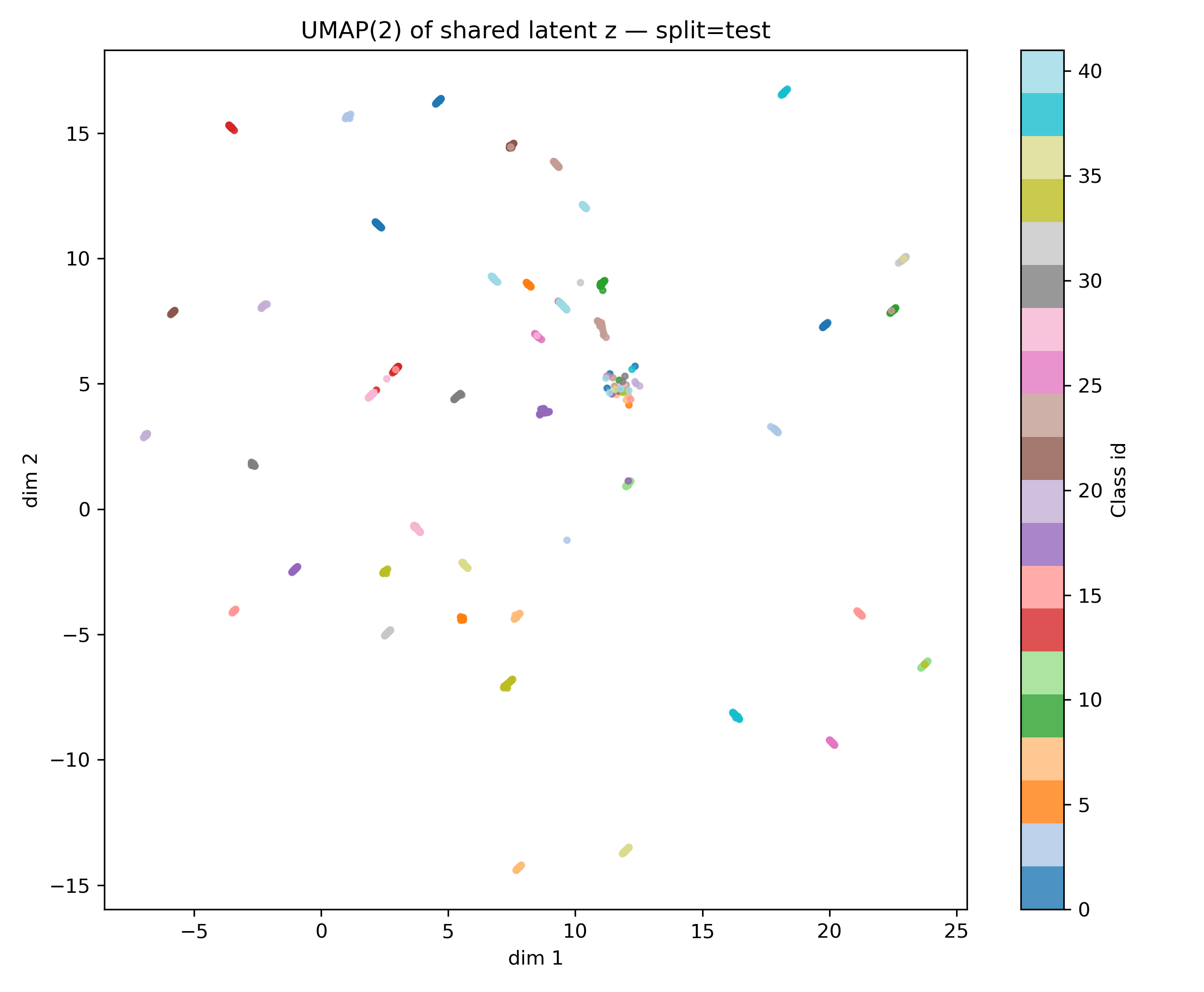}
    \caption{UMAP visualization confirming high cluster separability and smooth semantic transitions in the shared latent space.}
    \label{fig:umap_latent}
\end{figure}

\begin{figure}[htbp]
    \centering
    \includegraphics[width=0.48\textwidth]{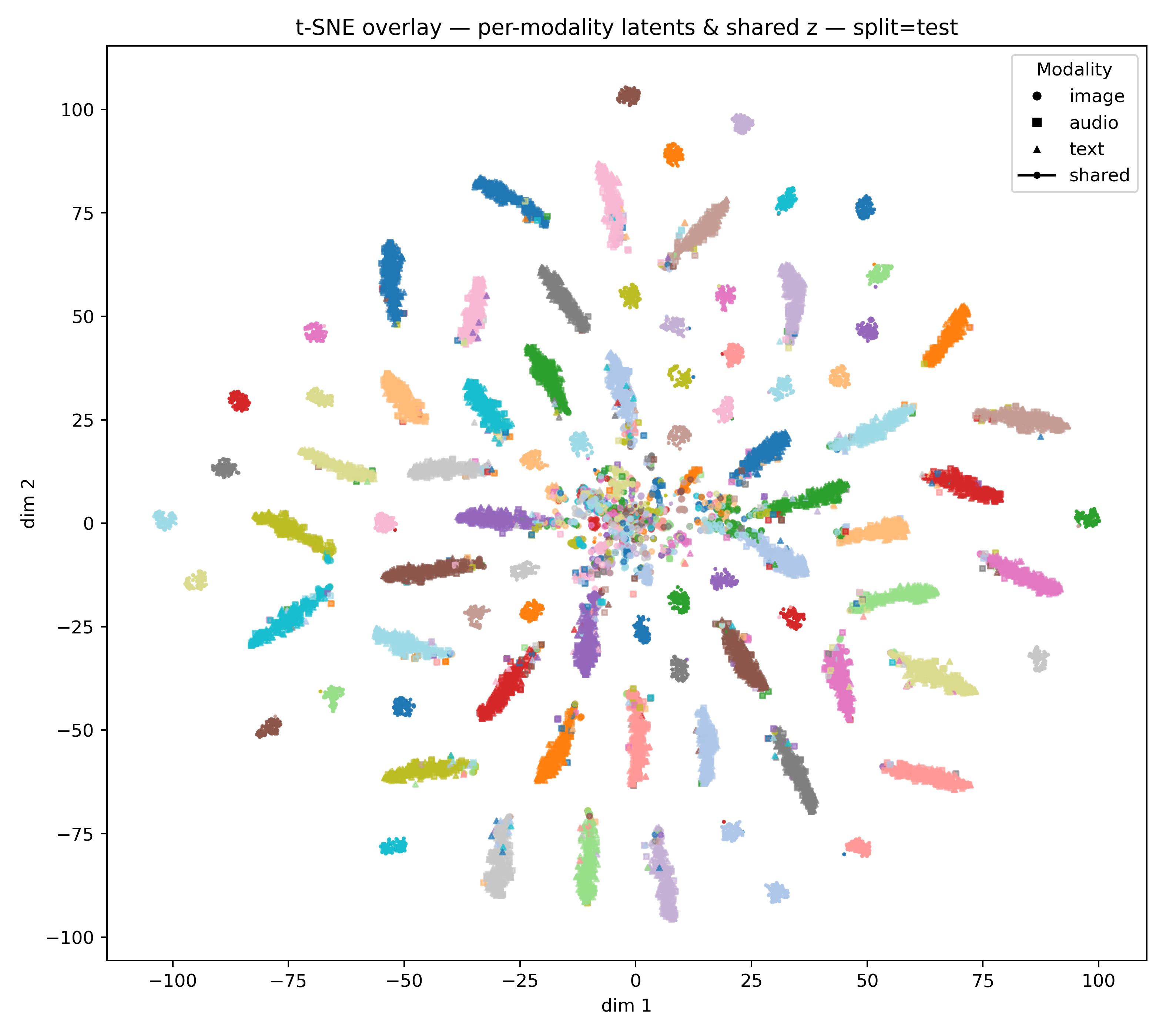}
    \caption{t-SNE overlay of latent embeddings by modality (image, audio, text). Convergence within shared clusters indicates modality invariance and cross-modal semantic consistency.}
    \label{fig:tsne_overlay}
\end{figure}

To visualize the learned multimodal embeddings, we applied t-SNE, PCA, and UMAP to both the baseline and MMAE representations. Figures~\ref{fig:tsne_latent} and~\ref{fig:umap_latent} show that the MMAE embeddings form well-separated clusters corresponding to semantic categories, while Figure~\ref{fig:tsne_overlay} demonstrates that samples from different modalities converge within the same cluster regions. This cross-modal overlap indicates that the model has successfully aligned semantically equivalent content—image, audio, and text—into a unified latent space.

Such behavior is particularly relevant for broadcast AI systems, where automated workflows often rely on mapping heterogeneous media elements (e.g., audio captions, visual frames, transcripts) to consistent metadata representations. The MMAE’s structure ensures that semantically similar items are embedded close together, facilitating downstream operations such as automated indexing, retrieval, and recommendation.

\subsection{Discussion}

The results confirm the central hypothesis that a reconstruction-driven approach yields more discriminative and semantically coherent multimodal embeddings than linear or unimodal methods. By enforcing joint reconstruction, the MMAE learns modality-invariant representations without the need for large-scale contrastive supervision, enabling efficient and interpretable learning from modest, well-aligned datasets like LUMA.

From an application standpoint, this finding is significant for broadcast and media systems. The learned latent representations can serve as unified semantic descriptors for media assets, supporting automated tagging, content-based retrieval, and cross-modal search. The MMAE’s lightweight architecture and data efficiency also make it deployable in production environments where compute resources or dataset sizes are limited.

Overall, the MMAE demonstrates that reconstruction-based multimodal representation learning offers a practical and data-efficient alternative to contrastive approaches such as CLIP and ALIGN. Its interpretability, robustness, and ability to generalize across modalities make it a promising foundation for next-generation AI-powered automation tools in broadcast content management, discovery, and archiving systems.

\section{Conclusion and Future Work}
\label{sec:conclusion}

This study demonstrated that reconstruction-driven multimodal learning can effectively uncover shared semantic representations across heterogeneous data sources. Using the LUMA dataset of aligned image–audio–text triplets, the proposed Multimodal Autoencoder (MMAE) outperformed linear fusion baselines in clustering quality and semantic coherence, confirming that joint reconstruction learning naturally aligns modalities within a shared embedding space.

Beyond improved performance, the MMAE offers scalability and interpretability advantages over contrastive methods like CLIP or ALIGN, achieving robust cross-modal alignment with limited, well-curated data in a fully unsupervised setting. This makes it especially suitable for media and broadcast contexts, where audiovisual and textual data are abundant but labeled datasets are scarce.

Practically, the MMAE’s shared latent space enables automation of key media workflows such as semantic tagging, cross-modal retrieval, and synchronized organization of transcripts, frames, and audio tracks. From an enterprise perspective, this framework provides a scalable and reproducible foundation for integrating multimodal AI into production environments—reducing manual labeling, enhancing content understanding, and accelerating intelligent media management at scale.

Future work will focus on extending the architecture with transformer-based encoders and probabilistic components for greater robustness to noise and missing modalities, as well as adapting the model to temporal data for dynamic video–audio–text synchronization and downstream tasks like caption generation and content classification.

In summary, the MMAE delivers a reproducible, explainable, and data-efficient approach to multimodal learning—bridging generative and contrastive paradigms while offering practical value for next-generation AI-driven automation in broadcast and enterprise media systems.

\bibliographystyle{IEEEtran}
\bibliography{references}

@inproceedings{radford2021clip,
  title={Learning Transferable Visual Models From Natural Language Supervision},
  author={Radford, Alec and Kim, Jong Wook and Hallacy, Chris and Ramesh, Aditya and Goh, Gabriel and Agarwal, Sandhini and Sastry, Girish and Askell, Amanda and Mishkin, Pamela and Clark, Jack and Krueger, Gretchen and Sutskever, Ilya},
  booktitle={Proceedings of the International Conference on Machine Learning (ICML)},
  year={2021}
}

@inproceedings{jia2021align,
  title={Scaling Up Visual and Vision-Language Representation Learning With Noisy Text Supervision},
  author={Jia, Chao and Yang, Yinfei and Xia, Ye and Chen, Yi-Ting and Parekh, Zarana and Pham, Hieu and Le, Quoc V. and Sung, Yunhsuan and Li, Zhen and Duerig, Tom},
  booktitle={Proceedings of the International Conference on Machine Learning (ICML)},
  year={2021}
}

@inproceedings{li2021albef,
  title={Align Before Fuse: Vision and Language Representation Learning with Momentum Distillation},
  author={Li, Junnan and Selvaraju, Ramprasaath R. and Gotmare, Akhilesh and Joty, Shafiq and Xiong, Caiming and Hoi, Steven},
  booktitle={Advances in Neural Information Processing Systems (NeurIPS)},
  year={2021}
}

@inproceedings{tan2019lxmert,
  title={LXMERT: Learning Cross-Modality Encoder Representations from Transformers},
  author={Tan, Hao and Bansal, Mohit},
  booktitle={Proceedings of the 2019 Conference on Empirical Methods in Natural Language Processing (EMNLP)},
  year={2019}
}

@inproceedings{ngiam2011multimodal,
  title={Multimodal Deep Learning},
  author={Ngiam, Jiquan and Khosla, Aditya and Kim, Mingyu and Nam, Jinyuan and Lee, Honglak and Ng, Andrew Y.},
  booktitle={Proceedings of the 28th International Conference on Machine Learning (ICML)},
  year={2011}
}

@article{suzuki2016jmvae,
  title={Joint Multimodal Learning with Deep Generative Models},
  author={Suzuki, Masahiro and Nakayama, Kotaro and Matsuo, Yutaka},
  journal={arXiv preprint arXiv:1611.01891},
  year={2016}
}

@inproceedings{wu2018mvae,
  title={Multimodal Generative Models for Scalable Weakly-Supervised Learning},
  author={Wu, Mike and Goodman, Noah},
  booktitle={Advances in Neural Information Processing Systems (NeurIPS)},
  year={2018}
}

@inproceedings{shi2019mmvae,
  title={Variational Mixture-of-Experts Autoencoders for Multi-Modal Deep Generative Models},
  author={Shi, Yuge and Paige, Brooks and Torr, Philip H. S.},
  booktitle={Advances in Neural Information Processing Systems (NeurIPS)},
  year={2019}
}

@article{suzuki2022survey,
  title={A Survey of Multimodal Deep Generative Models},
  author={Suzuki, Masahiro},
  journal={Advanced Robotics},
  volume={36},
  number={5-6},
  pages={261--278},
  year={2022},
  publisher={Taylor \& Francis}
}

@inproceedings{luma2024, series={SIGIR ’25},
   title={LUMA: A Benchmark Dataset for Learning from Uncertain and Multimodal Data},
   url={http://dx.doi.org/10.1145/3726302.3730302},
   DOI={10.1145/3726302.3730302},
   booktitle={Proceedings of the 48th International ACM SIGIR Conference on Research and Development in Information Retrieval},
   publisher={ACM},
   author={Bezirganyan, Grigor and Sellami, Sana and Berti-Équille, Laure and Fournier, Sébastien},
   year={2025},
   month=jul, pages={3782–3791},
   collection={SIGIR ’25} }

@inproceedings{li2022blip,
  title={BLIP: Bootstrapped Language-Image Pretraining for Unified Vision-Language Understanding and Generation},
  author={Li, Junnan and Li, Dongxu and Xiong, Caiming and Hoi, Steven},
  booktitle={Proceedings of the International Conference on Machine Learning (ICML)},
  year={2022}
}

@inproceedings{kong2020panns,
  title={PANNs: Large-Scale Pretrained Audio Neural Networks for Audio Pattern Recognition},
  author={Kong, Qiuqiang and Cao, Yin and Iqbal, Turab and Wang, Yong Xu and Plumbley, Mark D. and Wang, Wenwu},
  booktitle={IEEE/ACM Transactions on Audio, Speech, and Language Processing},
  year={2020}
}

@inproceedings{devlin2019bert,
  title={BERT: Pre-training of Deep Bidirectional Transformers for Language Understanding},
  author={Devlin, Jacob and Chang, Ming-Wei and Lee, Kenton and Toutanova, Kristina},
  booktitle={Proceedings of NAACL-HLT},
  year={2019}
}

@article{li2023blip2,
  title={BLIP-2: Bootstrapping Language-Image Pre-training with Frozen Image Encoders and Large Language Models},
  author={Li, Junnan and Lin, Ke and Gan, Zhe and others},
  journal={arXiv preprint arXiv:2301.12597},
  year={2023}
}

@inproceedings{he2022mae,
  title={Masked Autoencoders Are Scalable Vision Learners},
  author={He, Kaiming and Chen, Xinlei and Xie, Saining and Li, Yanghao and Doll{\'a}r, Piotr and Girshick, Ross},
  booktitle={IEEE/CVF Conference on Computer Vision and Pattern Recognition (CVPR)},
  year={2022}
}

@inproceedings{baevski2020wav2vec2,
  title={wav2vec 2.0: A Framework for Self-Supervised Learning of Speech Representations},
  author={Baevski, Alexei and Zhou, Yuhao and Mohamed, Abdelrahman and Auli, Michael},
  booktitle={Advances in Neural Information Processing Systems (NeurIPS)},
  year={2020}
}

@inproceedings{hsu2021hubert,
  title={HuBERT: Self-Supervised Speech Representation Learning by Masked Prediction of Hidden Units},
  author={Hsu, Wei-Ning and others},
  booktitle={Neural Information Processing Systems (NeurIPS)},
  year={2021}
}

@book{jolliffe2002pca,
  title={Principal Component Analysis},
  author={Jolliffe, Ian},
  publisher={Springer},
  year={2002}
}

@article{lloyd1982kmeans,
  title={Least Squares Quantization in PCM},
  author={Lloyd, Stuart},
  journal={IEEE Transactions on Information Theory},
  year={1982},
  volume={28},
  number={2},
  pages={129--137}
}

@article{hotelling1936cca,
  title={Relations Between Two Sets of Variates},
  author={Hotelling, Harold},
  journal={Biometrika},
  volume={28},
  pages={321--377},
  year={1936}
}

@article{tan2023multimodalreview,
  title={A comprehensive survey on multimodal learning: Fundamentals, challenges, and future directions},
  author={Tan, Hao and Zhang, Zhe and Wang, Xiaowei},
  journal={IEEE Transactions on Pattern Analysis and Machine Intelligence},
  year={2023},
  publisher={IEEE}
}

@article{jiang2022survey,
  title={Multimodal learning: A survey on foundations, methods and applications},
  author={Jiang, Mengmeng and Liu, Yuntao and Wang, Zhe and others},
  journal={IEEE Transactions on Neural Networks and Learning Systems},
  year={2022},
  publisher={IEEE}
}

@inproceedings{li2023mediaml,
  title={MediaML: Learning cross-modal embeddings for broadcast media analysis},
  author={Li, Yuchen and Gupta, Saurabh and Kuo, Cheng-Hao},
  booktitle={Proceedings of the 2023 IEEE International Conference on Multimedia and Expo (ICME)},
  year={2023},
  organization={IEEE}
}

@inproceedings{zhang2023crossmodal,
  title={Cross-modal representation learning for media retrieval and recommendation},
  author={Zhang, Lei and Xu, Ming and Yang, Fan},
  booktitle={Proceedings of the 2023 ACM Multimedia Conference},
  year={2023}
}

@article{chen2022visionlanguage,
  title={Vision-language pretraining: Current trends and future directions},
  author={Chen, Y. and Liu, P. and Zhao, S.},
  journal={ACM Computing Surveys},
  year={2022}
}

@inproceedings{gupta2023broadcastai,
  title={Broadcast AI: Towards intelligent multimodal content indexing and retrieval},
  author={Gupta, Arjun and Ramesh, Priya and Singh, Rahul},
  booktitle={Proceedings of the 2023 NAB Broadcast Engineering and IT Conference},
  year={2023}
}

@inproceedings{gao2022unified,
  title={Unified multimodal representation learning for content understanding},
  author={Gao, Tianyu and Sun, Wei and Huang, Lei},
  booktitle={NeurIPS 2022 Workshop on Multimodal Learning},
  year={2022}
}

@article{xu2023survey,
  title={A survey on multimodal representation learning: From alignment to fusion},
  author={Xu, Lin and He, Y. and Wang, T.},
  journal={IEEE Transactions on Multimedia},
  year={2023}
}

@inproceedings{muller2023audiofusion,
  title={AudioFusion: Cross-modal alignment of audio and visual signals for broadcast applications},
  author={Muller, Julian and Zhao, Rui and Ahmed, Samir},
  booktitle={IEEE ICASSP 2023},
  year={2023}
}

@article{zhao2022deepmultimodal,
  title={Deep multimodal learning: Methods, applications, and challenges},
  author={Zhao, Qiang and Liu, Ning and Zhang, Rui},
  journal={Pattern Recognition},
  year={2022}
}

@inproceedings{huang2023robustfusion,
  title={Robust multimodal fusion under noisy and missing data},
  author={Huang, Jie and Xu, Kai and Luo, Weijia},
  booktitle={Proceedings of the IEEE/CVF Conference on Computer Vision and Pattern Recognition (CVPR)},
  year={2023}
}

@inproceedings{shen2023lightclip,
  title={LightCLIP: A lightweight multimodal framework for data-efficient training},
  author={Shen, Tao and Yu, Jing and Peng, Hao},
  booktitle={Proceedings of the 2023 Conference on Empirical Methods in Natural Language Processing (EMNLP)},
  year={2023}
}

@article{zhang2023efficientmultimodal,
  title={Efficient multimodal transformers: A survey and future directions},
  author={Zhang, Bo and Wang, Cheng and Liu, Wen},
  journal={Information Fusion},
  year={2023}
}

@article{fang2023survey,
  title={Contrastive multimodal learning: A survey of methods and applications},
  author={Fang, Xin and He, Chen and Li, Wen},
  journal={IEEE Access},
  year={2023}
}

@article{li2023foundational,
  title={Foundational multimodal models: Unifying vision, language, and audio understanding},
  author={Li, Rui and Zhang, Kai and Wang, Dong},
  journal={Artificial Intelligence Review},
  year={2023}
}

@inproceedings{han2023mediaclip,
  title={MediaCLIP: Bridging vision-language models with broadcast media datasets},
  author={Han, Xiaoyu and Wang, Rui and Li, Qian},
  booktitle={Proceedings of the IEEE International Conference on Image Processing (ICIP)},
  year={2023}
}

@article{rao2024multimodalvision,
  title={Multimodal vision systems: Foundations, advances, and applications},
  author={Rao, Di and Liu, Qing and Zhao, Fei},
  journal={IEEE Transactions on Multimedia},
  year={2024}
}

@article{guo2023multimodalvae,
  title={Multimodal variational autoencoders for self-supervised representation learning},
  author={Guo, Lin and Wu, Hao and Sun, Chen},
  journal={Neural Networks},
  year={2023}
}

@article{wu2022multimodalautoencoder,
  title={Multimodal autoencoders: A unified architecture for cross-modal representation learning},
  author={Wu, Tian and Li, Wen and Chen, Bo},
  journal={IEEE Transactions on Neural Networks and Learning Systems},
  year={2022}
}

@article{zhou2024broadcastai,
  title={Broadcast AI 2.0: Intelligent automation and multimodal understanding for next-generation media systems},
  author={Zhou, Wei and Park, Jiyoon and Cheng, Hao},
  journal={IEEE Transactions on Broadcasting},
  year={2024}
}

@article{wang2023multimodalframework,
  title={A lightweight multimodal framework for media intelligence and broadcast content analysis},
  author={Wang, Kai and Li, Tian and Zhao, Yujing},
  journal={Journal of Broadcasting and Electronic Media},
  year={2023}
}

@inproceedings{wang2023unimodal,
  title={UniModal: Unified multimodal alignment via joint representation reconstruction},
  author={Wang, Yonghui and Chen, Jie and Xu, Bo},
  booktitle={NeurIPS 2023},
  year={2023}
}

@inproceedings{xu2022bridgetower,
  title={BridgeTower: Building bridges between vision and language representations for unified multimodal learning},
  author={Xu, Pengchuan and Wang, Xin and Li, Qi},
  booktitle={EMNLP 2022},
  year={2022}
}

@inproceedings{akbari2021vatt,
  title={VATT: Transformers for multimodal self-supervised learning from raw video, audio and text},
  author={Akbari, Hassan and Yuan, Li and Recasens, Adriana and others},
  booktitle={NeurIPS 2021},
  year={2021}
}

@inproceedings{rombach2022latentdiffusion,
  title={High-resolution image synthesis with latent diffusion models},
  author={Rombach, Robin and Blattmann, Andreas and Lorenz, Dominik and others},
  booktitle={Proceedings of the IEEE/CVF Conference on Computer Vision and Pattern Recognition (CVPR)},
  year={2022}
}

@inproceedings{bao2022beit3,
  title={BEiT-3: Multi-modal pretraining for unified vision-language understanding and generation},
  author={Bao, Hangbo and Wang, Wenhui and Dong, Li and others},
  booktitle={Proceedings of the 2022 European Conference on Computer Vision (ECCV)},
  year={2022}
}

\end{document}